\documentclass{article}
\usepackage{graphicx}

\begin{document}

\title{Optimization of Supply Diversity for the Self-Assembly of Simple Objects
in Two and Three Dimensions}

\author{Fabio R. J. Vieira\\
Valmir C. Barbosa\thanks{Corresponding author (valmir@cos.ufrj.br).}\\
\\
Universidade Federal do Rio de Janeiro\\
Programa de Engenharia de Sistemas e Computa\c c\~ao, COPPE\\
Caixa Postal 68511\\
21941-972 Rio de Janeiro - RJ, Brazil}

\date{}

\maketitle

\begin{abstract}
The field of algorithmic self-assembly is concerned with the design and analysis
of self-assembly systems from a computational perspective, that is, from the
perspective of mathematical problems whose study may give insight into the
natural processes through which elementary objects self-assemble into more
complex ones. One of the main problems of algorithmic self-assembly is the
minimum tile set problem (MTSP), which asks for a collection of types of
elementary objects (called tiles) to be found for the self-assembly of an object
having a pre-established shape. Such a collection is to be as concise as
possible, thus minimizing supply diversity, while satisfying a set of stringent
constraints having to do with the termination and other properties of the
self-assembly process from its tile types. We present a study of what we think
is the first practical approach to MTSP. Our study starts with the introduction
of an evolutionary heuristic to tackle MTSP and includes results from extensive
experimentation with the heuristic on the self-assembly of simple objects in two
and three dimensions. The heuristic we introduce combines classic elements from
the field of evolutionary computation with a problem-specific variant of
Pareto dominance into a multi-objective approach to MTSP.

\end{abstract}

\section{Introduction}\label{sec:intro}

Self-assembly is the process whereby simple building blocks self-organize into
more complex structures. Its occurrence in nature is ubiquitous and can be
observed at various scales, ranging from the growth of crystals, through the
production of complex molecules inside the living cell, to the formation of
galaxies and even their clustering into larger structures. In recent years, a
considerable amount of research has been directed toward attempting to
understand and reproduce the essential mechanisms that drive self-assembly,
aiming at amassing its power of distributed control for goal-directed building
tasks of strategic importance \cite{r02,mummtbbrdsdrgsslg07}. The premise has
been that, if successful, such efforts may eventually lead to autonomous systems
that, for example, will replace the current process of photolithography in VLSI
fabrication \cite{pld06}, or the current electronic media for information
storage \cite{r06}, or yet will give rise to cooperating teams of robots for
nano- and large-scale construction through rules of self-assembly
\cite{tgtmbd06,wbrn06} or to solvers of hard mathematical problems through DNA
computing \cite{w96,bcm04,b07}. At the current stage, a lot of research is being
directed toward the biochemical production of nanoscale building blocks
\cite{wlws98,ynlpbpgrhc05,r06,ra06,pd07} or devices \cite{pld06,r06,s07}, or the
elaboration of logical rules for the programming of robots \cite{tm04,wbrn06}.

The mathematical study of self-assembly can be said to have begun as far back as
\cite{w61} on the formation of carbon crystals from the nanoscale structures
that the author called tiles. However, it seems to have been only much more
recently, after algorithm- and complexity-related notions were sufficiently
mature from their development within computer science, that momentum began to
accumulate. The resulting discipline is currently known as algorithmic
self-assembly. It is based on the two-dimensional model laid down in
\cite{w98,rw00}, itself an extension of the earlier, one-dimensional model of
\cite{acghker02}. The model is built on an unbounded two-dimensional grid at
whose nodes square tiles can be placed so that tiles that occupy neighboring
nodes are themselves juxtaposed with a side in common. Each tile is labeled on
at least one of its four sides and to each label there corresponds a unique
positive integer indicating the intensity that the bond formed by juxtaposing
two tiles on sides of equal labels will have. Tiles may occur in a finite number
of types, each capable of supplying an unbounded number of identical tiles.

In this model, the process of self-assembly begins with the placement of a
special tile, called the seed, at an arbitrary node. It then proceeds in
discrete time steps by accreting one tile of one of the available types per
time step. It is important to note that growth can only take place by accretion,
that is, by aggregation to the exterior of the current object, which expressly
prevents hollows that are left from being filled later on. The model also
includes a positive integer parameter, called a temperature, intended to
regulate the addition of new tiles. Specifically, a new tile may only be added
if bonds are formed only on sides of the same label, and furthermore the net
intensity of the resulting (at most three) bonds surpasses the temperature. A
sequence of two-dimensional objects is then established of which the first
comprises the seed tile only and each of the others is the augmentation of its
precursor by the accretion of exactly one tile.

A considerable body of knowledge has been developed on this model, targeting
primarily the establishment of theoretical properties of the objects produced
from both a structural and a functional perspective
\cite{ccghe04,acgkes05,rsy05,akv06,w06}. Along with these properties,
optimization problems have been defined that embody the essential difficulties
associated with translating all the theoretical discoveries into the practical
context that motivated the whole field in the first place. One of these problems
is that of finding out how diverse the supply of tiles for self-assembly has to
be in order for an object of pre-established shape to be produced. In other
words, the problem asks for the minimum number of tile types for assembling an
object of the desired shape.

This problem is known as the minimum tile set problem (henceforth, MTSP)
\cite{acghker02}. Given a temperature and a two-dimensional shape, MTSP asks for
a seed tile and the least diverse set of tile types out of which an object of
the desired shape may be assembled. MTSP is constrained by the requirements that
all object sequences obtained from the seed terminate, that they all terminate
in an object of the desired shape, and furthermore that the objects obtained at
termination have the property of being full (in the sense of containing as many
bonds as there can be juxtaposed square sides). Formally, MTSP is an NP-hard
problem \cite{acghker02}. While to the practitioner this is generally an
indication of inherent computational difficulty and a clear sign that heuristics
are to be sought to solve the problem, in the case of MTSP even choosing an
adequate heuristic has the feel of something arduous, owing in essence to the
fact that the problem's constraints (termination, unicity, and fullness) seem to
loom too large upon the computational resources one can usually count on.

Our subject in this paper is the development of a heuristic to tackle MTSP. The
heuristic we describe is an evolutionary algorithm for multi-objective
optimization: it uses a convenient variation of Pareto dominance along with
suitable crossover and mutation operators to drive the search in the direction
of concise tile sets that do nonetheless satisfy the problem's constraints. We
report on experiments for the two-dimensional model discussed above, for a
variation thereof that allows tiles to be rotated on the plane before being
accreted onto the growing object, and likewise for the three-dimensional
extension of the two-dimensional model with rotation. These two additional
models appear, to our knowledge, nowhere else in the literature on algorithmic
self-assembly. They seem, however, to enable more realistic representations of
real-world processes of self-assembly, as argued in \cite{ho06} in the case of
three dimensions, and have for this reason been included.

We are aware of no other study targeting a practical approach to
MTSP.\footnote{The work reported in \cite{tkkg05} is the only one to come
somewhat close to being an exception, but its relation to our work remains very
tenuous, since the problem that is handled there does not share the objective or
constraints of MTSP, even though it too targets the determination of tile
types.} While this is surprising per se, in the context of the present paper it
has also hindered the possibilities for a comparative assessment of our
heuristic vis-\`a-vis others severely. On the computational side, then, we have
concentrated on squares and cubes as the desired final shapes. For the original
two-dimensional model, at least, this has given us a solid basis for comparison,
since upper bounds on the number of necessary tile types are available from
\cite{rw00}. For the other two models no such knowledge is available, but we
expect relatively smaller tile sets to suffice when the possibility of rotation
is added to the two-dimensional model, which then gives us an indirect basis on
which to judge what we find in this case. As for the three-dimensional case, all
we can achieve at this point is a qualitative evaluation with respect to the
two-dimensional cases.

We proceed in the following manner. We first detail MTSP and also the three
models we use in Section~\ref{sec:mtsp}. Then we present our heuristic with all
its elements in Section~\ref{sec:heur} and computational results in
Section~\ref{sec:results}. Further comments and concluding remarks are given in
Section~\ref{sec:concl}.

\section{The Minimum Tile Set Problem}\label{sec:mtsp}

\subsection{Problem Formulation}

The formulation of MTSP we give in this section is good for the two-dimensional
models and for the three-dimensional model. We use the word tile to refer both
to a square and to a cube, and likewise the word side to refer both to a side of
a square and a face of a cube. We do this for conciseness and no confusion
should arise.

Let $\mathcal{T}$ denote a set of tile types and let $S$ be the seed tile. The
process of self-assembly from $S$ and $\mathcal{T}$ assumes the existence of an
unbounded supply of tiles of all types in $\mathcal{T}$ and is represented by a
sequence $O=\langle\omega_1,\omega_2,\ldots\rangle$, where $\omega_1$ is the
object constituted by the single tile $S$. For $u>1$, $\omega_u$ is the object
obtained from $\omega_{u-1}$ by the accretion of exactly one tile of one of the
types in $\mathcal{T}$ for which model-dependent binding conditions are
satisfied. Such conditions will be specified shortly for each model, but they
all have in common the property that, as the new tile is accreted onto
$\omega_{u-1}$, a number of bonds is created that is at least one and at most
the number of sides in the tile through which binding can occur (provided at
least one of the tile's sides is left unbound, given that growth is effected by
accretion). If $t$ is a tile in the resulting $\omega_u$, we let $b_u(t)$ denote
the number of bonds involving $t$ in $\omega_u$. Similarly, we let $B_u$ be the
total number of bonds holding $\omega_u$ together. Clearly,
$B_u=\sum_tb_u(t)/2$, where the summation ranges over all tiles in $\omega_u$.

Now let $\Omega_S$ be the set of all sequences that represent a self-assembly
process from $S$. Also, let $\omega^*$ be any object having the desired,
pre-specified shape and let $B^*$ be the maximum number of bonds that may exist
in an object of that shape. If $\tau>0$ is the temperature, then MTSP requires
that $S$ and $\mathcal{T}$ be determined such that $\vert\mathcal{T}\vert$ is
minimum over all seeds and tile-type sets that satisfy the following
constraints:
\begin{enumerate}
\item[C1.] (Termination) For all $O\in\Omega_S$, $O$ is finite; let
$\vert O\vert$ be the number of objects in $O$.
\item[C2.] (Unicity) For all $O\in\Omega_S$, $\omega_{\vert O\vert}$ has the
same shape as $\omega^*$.
\item[C3.] (Fullness) For all $O\in\Omega_S$, $B_{\vert O\vert}=B^*$.
\end{enumerate}

Notice, in this formulation, that the temperature $\tau$ is never mentioned
explicitly. This is because the role it plays is in helping specify the binding
conditions that allow a new tile to be accreted onto an object. These conditions
are given below as part of each model's characteristics, but the way they relate
to $\tau$ is common to all three models. We handle this before discussing the
models.

Each tile type $T\in\mathcal{T}$ is characterized by a number $\sigma(T)$ of
sides (either $4$ or $6$, respectively in two and three dimensions) and by a
number $\lambda(T)$ of sides through which binding is possible. These numbers
are such that $1\le\lambda(T)\le\sigma(T)$. Each of the $\lambda(T)$ sides is
labeled to indicate how binding through that side does actually take place. If
$\ell$ is one such label, then whenever binding occurs through a side that is
thus labeled the resulting bond has intensity given by the integer $I(\ell)>0$.
The $\tau$-dependent condition for a tile $t$ of type $T$ to be accreted onto an
object at a certain place is that
$I(\ell_1)+\cdots+I(\ell_{\lambda(t)})\ge\tau$, where
$\ell_1,\ldots,\ell_{\lambda(t)}$ are the labels on the sides through which
binding does take place (and therefore $\lambda(t)\le\lambda(T))$. What is left
to specify of each model's binding conditions is precisely the $\lambda(t)$
sides.

\subsection{Two-Dimensional Model (2D)}

\begin{figure}
\centering
\includegraphics[scale=0.30]{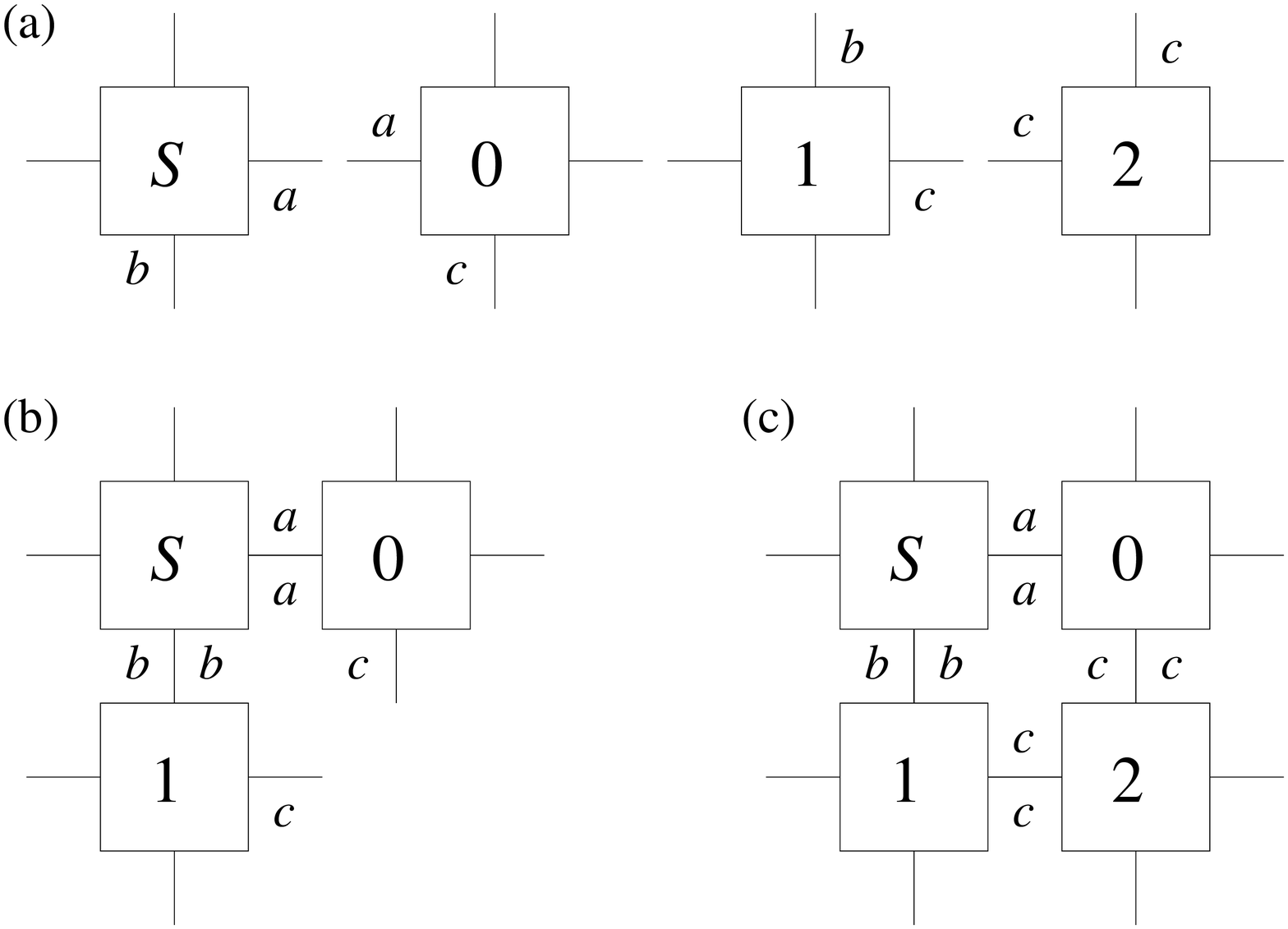}
\caption{In the 2D model, using the tile types in (a) allows the object in (c)
to be obtained from the object in (b) by accreting a tile of type $2$ onto it.
Temperature is $\tau=2$. Label intensities are $I(a)=I(b)=2$ and $I(c)=1$.}
\label{fig:2dexample}
\end{figure}

This is the original model of \cite{rw00} and is characterized by a fixed
orientation of all tile types with respect to the underlying two-dimensional
lattice. That is, if we identify ``north,'' ``east,'' ``south,'' and ``west''
directions on the lattice and proceed likewise on all tile types, then a tile
may only be accreted onto a growing object if its directions are aligned with
those of the lattice. This given, the binding conditions for this model require
identical labels on the sides that face each other for a bond to be created
between them. Thus, accreting a tile $t$ onto an object requires
$\lambda(t)\le 3$ pairs of matched labels. This is illustrated in
Figure~\ref{fig:2dexample}.\footnote{In this figure, and in others to follow for
the two-dimensional cases, juxtaposed tiles are represented with a gap between
them so that the matching labels can be shown explicitly whenever possible.}

The 2D model may seem too restrictive or even implausible, but from \cite{rw00}
we know what seems to be tight upper bounds on the value of the optimal
$\vert\mathcal{T}\vert$ when the desired shape is that of an $n\times n$ square
with $n\ge 3$. The 2D model is then invaluable in our present context, as it
provides the only available benchmark with which the results we obtain can be
compared. What is demonstrated in \cite{rw00} is that, for $\tau=1$, and letting
$\mathcal{T}^*$ be an optimal tile set, $\vert\mathcal{T}^*\vert\le n^2$. For
$\tau=2$,
\begin{equation}
\vert\mathcal{T}^*\vert\le\cases{
n+4,&if $3\le n\le 23$;\cr
22+\lceil\log_2 n\rceil,&if $23\le n<22+2^{23}$;\cr
22\log_2^*n,&if $22+2^{23}\le n$\cr
}
\label{eq:upperbound}
\end{equation}
($\log_2^*$ indicates the number of times one must apply the $\log_2$ operator
in succession until the result is no greater than $1$). As for $\tau\ge 3$, it
is unknown whether the bounds in (\ref{eq:upperbound}) can be improved.

\subsection{Two-Dimensional Model with Rotation (2DR)}

\begin{figure}
\centering
\includegraphics[scale=0.30]{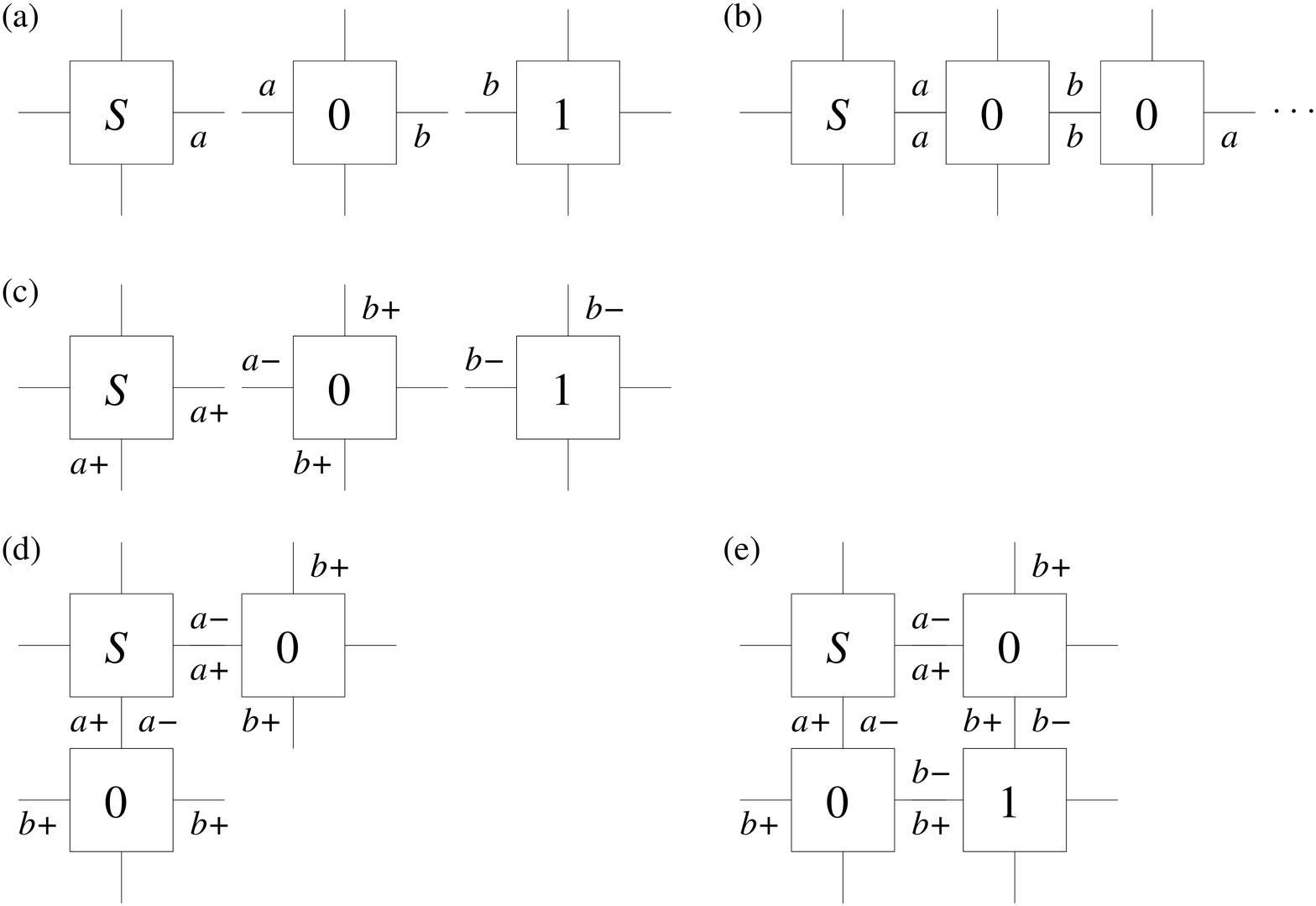}
\caption{In the 2DR model, using the tile types in (a) may lead to the
unbounded growth shown in (b). Using the tile types in (c), where polarities are
also used, allows the object in (e) to be obtained from the object in (d) by
accreting a tile of type $1$ onto it. Temperature is $\tau=2$. Label intensities
are $I(a)=I(b)=2$ in (a) and (b), $I(a)=2$ and $I(b)=1$ in (c)--(e).}
\label{fig:2drexample}
\end{figure}

Our main motivation to introduce the 2DR model has been the possibility of
increased plausibility that comes with allowing clockwise or counterclockwise
rotation of a tile by multiples of $90^\circ$ before accreting it onto an
object.\footnote{Notice that this is the only kind of rotation that makes sense
in two dimensions. Possibilities like flipping a tile around its north-south
or east-west axis, for example, implicitly require incursions into the third
dimension and must not be considered.} Rotations are intrinsically problematic,
though: as shown in Figure~\ref{fig:2drexample}, there exist tile-type sets
that, by virtue of the added possibility of rotation, make it impossible for
constraint C1 to be satisfied. We circumvent this problem by adopting the
binding conditions of the 2D model and making them more stringent. The idea is
to let each label have a polarity (either $+$ or $-$) and to require different
polarities, in addition to equal labels, for the bond to be created. Similarly
to the 2D model, accreting a tile $t$ onto an object requires $\lambda(t)\le 3$
pairs of matched labels, each pair with different polarities. This is also
illustrated in Figure~\ref{fig:2drexample}.

Unlike the 2D model, there are now no known upper bounds on
$\vert\mathcal{T}^*\vert$. But the 2DR model is more flexible than the 2D model,
in the sense that, in principle, the same tile type may be used in different
situations, depending on how tiles are rotated. In fact, each tile type in 2DR
stands for up to $4$ distinct tile types in 2D, so we expect solutions to be no
worse than in the 2D case, possibly even better.

\subsection{Three-Dimensional Model with Rotation (3DR)}

\begin{figure}
\centering
\includegraphics[scale=0.50]{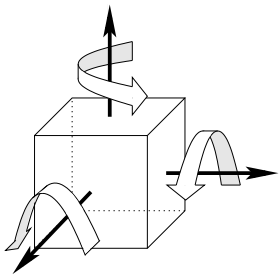}
\caption{A tile's possible rotations in the 3DR model.}
\label{fig:3drotations}
\end{figure}

As we move into three dimensions, motivation is derived directly from the field
of molecular dynamics \cite{ch03} and, in the absence of any upper bounds like
those given in (\ref{eq:upperbound}) for the 2D model, it is very hard to
justify the impossibility of rotation. We have then skipped what would be a 3D
model and moved directly to 3DR, three-dimensional with the possibility of
rotation.  The underlying lattice is now three-dimensional and the possible
rotations are by multiples of $90^\circ$ around one of the three possible axes,
as shown in Figure~\ref{fig:3drotations}. As a consequence, each tile type in
3DR stands for up to $24$ distinct tile types in the (hypothetical) 3D model.

\begin{figure}
\centering
\includegraphics[scale=0.30]{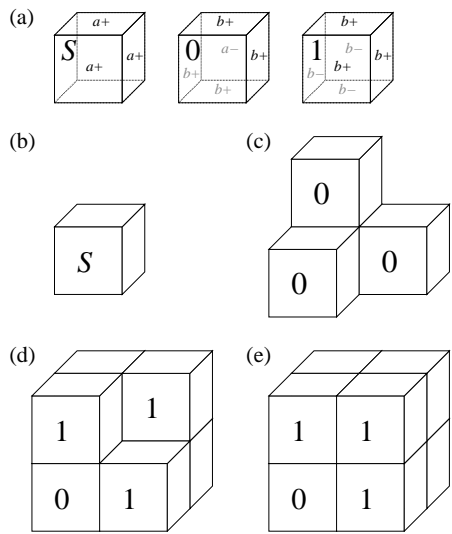}
\caption{In the 3DR model, using the tile types in (a) allows the object in (e)
to be obtained from the object in (b) through the following sequence of
accretions: first three tiles of type $0$ are accreted, one at a time, onto the
growing object that results in the one in (c); then three tiles of type $1$ are
used to lead to the object in (d); then a fourth object of type $1$ is used.
Temperature is $\tau=2$. Label intensities are $I(a)=2$ and $I(b)=1$.}
\label{fig:3drexample}
\end{figure}

Rotations are just as problematic in 3DR as they are in 2DR, and for the same
reason. We then continue to use label polarity, so that accreting a tile $t$
onto an object now requires $\lambda(t)\le 5$ matching label pairs, each with
different polarities, as illustrated in Figure~\ref{fig:3drexample}. The 3DR
model also shares with the 2DR model the absence of benchmarks against which to
experiment, but as we move from square shapes in two dimensions to cubic shapes
in three dimensions an interesting trade-off turns up. On the one hand, the
added third dimension allows for more complex shapes to be considered; on the
other hand, rotating tiles have more degrees of freedom in three dimensions than
in two. Just which trend may eventually have the upper hand is unclear now, but
we are given some basis for evaluation in three dimensions, at least in
qualitative terms.

\section{An Evolutionary Heuristic}\label{sec:heur}

\subsection{Overview}

Our heuristic is evolutionary and follows one of the customary layouts of a
generational genetic algorithm running on a population of tile-type sets, each
possibly of a different size. It runs for a fixed number of generations, each
corresponding to a population of fixed size, the first population created
randomly. Each of the subsequent populations is obtained from its precursor by
first an elitist step that seeks to preserve some of the fittest individuals,
then a diversity-preserving step that samples the previous population uniformly
at random and moves the resulting individuals to the new one, and then a step
that makes use of crossover and mutation and is repeated until the new
population is full. The diversity-preserving step never chooses from those
individuals already picked by the elitist step.

At each repetition of the crossover- and mutation-based step, a choice is made
as to whether the crossover or the mutation operator will be applied. In the
former case, two individuals are selected and combined, and the resulting two
individuals are added to the new population. In the latter case, one single
individual is selected and mutated before being added to the new population. The
choice is probabilistic and we have found that increasing the probability of
choosing crossover from generation to generation sometimes provides the best
results. If $G$ is the number of generations and $p_g$ is the probability of
choosing crossover in the $g$th generation, $1\le g\le G$, then we use
\begin{equation}
p_g=p_1+(g-1)\left(\frac{p_G-p_1}{G-1}\right),
\label{eq:xprob}
\end{equation}
that is, we let the crossover probability increase linearly from the initial
value $p_1$ through the final value $p_G$.

We do fitness-based selection, both in the elitist step and in each of the steps
that choose between crossover and mutation, based on a multi-objective approach.
While selecting from the population of generation $g$, we first group the
individuals into layers of Pareto-like dominance, then we choose one of the
layers probabilistically in a linearly normalized fashion, and then we select an
individual from the chosen layer uniformly at random. If $L_g$ denotes the
number of layers at this generation and $W_g$ the factor by which the top layer
(the most dominant, assumed to be layer $1$) is more likely than the bottom
layer (the least dominant, assumed to be layer $L_g$) to be chosen, then the
probability that the $l$th layer is chosen, $1\le l\le L_g$, is proportional to
$W_g-(l-1)(W_g-1)/(L_g-1)$. We have also found that the best results often come
from letting $W_g$ increase from generation to generation, since in the earliest
generations the fitness components vary little from one individual to another
and the layering process is practically devoid of meaning. As in the case of the
crossover probability, we let it increase linearly from the initial value $W_1$
through the final value $W_G$, that is,
\begin{equation}
W_g=W_1+(g-1)\left(\frac{W_G-W_1}{G-1}\right).
\label{eq:wnorm}
\end{equation}

\subsection{The Individual and its Simulation}

Each individual represents a set of tile types whose size lies somewhere between
a minimum and a maximum value, these bounds being the same for all individuals.
Each tile type is characterized by a set of labels and, for each label, an
intensity. During the evolutionary process there are two occasions in which
labels must be selected randomly from some universe of possibilities: the first
of these is the formation of the initial population; the second is when
individuals are to be mutated. For use in these occasions, a fixed table of
labels and associated intensities is used. Given the temperature $\tau$, this
table includes a certain variety of labels for each of the possible intensities
$1,\ldots,\tau$. It also includes the special label $\epsilon$, used to denote
the absence of a label. Just how much variety is included depends on the size of
the table, which is fixed beforehand. Given this size, the actual table is
obtained by truncating the sequence of distinct labels
$\epsilon,\ell_1,\ldots,\ell_\tau,\ell_{\tau+1},\ldots,\ell_{2\tau},\ldots$ to
as many labels as that desired size. As for the intensities of the
non-$\epsilon$ labels, we let
$I(\ell_{k\tau+1})=1,\ldots,I(\ell_{(k+1)\tau})=\tau$ for $k=0,1,\ldots$. Labels
in this table have no polarities: if in the 2DR or 3DR model, polarities are
chosen at random after each label has itself been chosen.

In order for an individual to be evaluated, we need to find out how close it
comes to allowing an object of the desired shape to be assembled, and also how
close it comes to satisfying the remainder of constraints C1 through C3. The
first step toward these goals is to simulate the process of self-assembly for
the individual in question, sometimes more than once, since it is unavoidable
that each simulation be essentially of a stochastic nature. The results of the
simulations can then be fed into appropriate fitness components, which in turn
can be used to perform the desired evaluation.

Each simulation is carried out on a lattice with periodic boundaries having the
same size along each dimension.\footnote{Periodic boundaries are such that
traversing the lattice along any dimension necessarily leads to an already
visited node.} The use of such boundaries allows the seed $S$ to be placed
arbitrarily on the lattice, which is necessarily finite for the purposes of
simulation. With $S$ in place, the simulation proceeds by maintaining two lists
and managing the interactions between them. At all times, the first list
contains all the lattice positions at which it is currently possible to add a
new tile to the growing object without making it collide with itself along
one of the dimensions. This list includes positions inside hollows, but this is
not detrimental to the notion of growth by accretion, as we discuss shortly.
The second list, in turn, contains all tiles that can currently be added to at
least one of the available lattice positions. Each tile type of the individual
under consideration may be represented in this list by several tiles, one for
each of the possible lattice positions at which it can be placed and for each
possible rotation (if 2DR or 3DR is the model in use). All that is required of
each such tile is that it satisfy the binding conditions for the model under
which the simulation is being conducted and also that all pertinent intensities
add up to at least $\tau$. Also, if the lattice position to which the tile
corresponds is inside a hollow, all these requirements must not involve a
necessary bond with the tile that was the one to close the hollow.

The simulation repeatedly selects a matching pair of elements from the two
lists, places the corresponding tile at the corresponding lattice position, and
updates the lists. The selection is probabilistic and is done proportionally to
the added intensities of the bonds to be created by each pair. This is done
until the tile list becomes empty or a pre-specified maximum number of tiles has
been used. If exit occurs on the former condition and the possibility of a
collision never occurred, then the simulation is said to have reached a terminal
object (this is not to say that the simulation would not terminate if it were
allowed to continue past the maximum number of tiles, or if the lattice were
infinite and collisions impossible, but merely reflects an arbitrary decision
aimed at keeping the simulation bounded in both time and space).

It is important to note that, when a tile is placed inside a hollow during the
simulation, this by no means indicates that the requirement of growth by
accretion only has been given up. Because the bond, if any, created between such
a tile and the one that closed the hollow earlier in the simulation is
inessential (i.e., the tile could be placed in the hollow and satisfy all
requirements even without such a bond), it is possible to accrete both tiles
onto the object at the same positions but in reversed order and still obtain the
same final result. The possibility of placing tiles inside hollows during
simulation is then only a device to help reorder accretive additions to the
growing object when they happen to be selected in an unfavorable order.

The simulation is repeated until a terminal object is reached or else a
pre-specified number of simulations has been performed. In the former case, the
results to be fed into the fitness components are obtained from the terminal
object that was reached. In the latter case, they are obtained from the object
with fewest tiles reached as the simulations ended. In either case, then, a
single object is output by the simulations.

Before discussing what the results of interest are for the individual at hand,
it is important to recall from the definition of MTSP that $S$, the seed, is
part of what has to be determined. It would then seem like some strange
dependency exists as far as the simulation is concerned, since it starts
precisely with placing $S$ on the lattice while, to comply with the definition
of MTSP, $S$ should be part of the simulation's output. What we do to avoid
this is to assume, initially, that $S$ is some sort of universal seed, that is,
a tile with wildcard labels on some of its sides (and $\epsilon$ on the others),
each of which ``becomes'' the appropriate label of whatever other tile is
accreted onto $S$ through it. At the end of the simulation, $S$ is the tile
obtained from the original, universal $S$ by substituting the label $\epsilon$
for all wildcard labels that remain.

Let $i$ denote the individual under consideration and $\omega(i)$ the object
output by the simulations on $i$. The simulations' results can be summarized as
follows:
\begin{itemize}
\item The number of tile types actually used in assembling $\omega(i)$, denoted
by $\theta(i)$. We have $\theta(i)\ge 0$, with $\theta(i)=0$ corresponding to
the case in which $\omega(i)$ comprises $S$ only.
\item The number of tiles that constitute $\omega(i)$, denoted by
$\vert\omega(i)\vert$. We have $\omega(i)\ge 1$.
\item The maximum number of tiles that $\omega(i)$ and $\omega^*$ (the object
having the desired shape) may have in common under all possible superpositions
of the two, including object rotations. The resulting number is denoted by
$\kappa(i)$. We have $\kappa(i)\ge 1$, with $\kappa(i)=1$ corresponding to the
case in which $\omega(i)$ comprises $S$ only or $\omega^*$ is one single tile.
\item The number of alternative tile types that could have been used at each
step of the construction of $\omega(i)$ from the $S$ that resulted at the end of
the simulation, all else unchanged up to that step, provided $\theta(i)>0$. In
order to be an alternative tile type at step $u$, a tile type must be one of the
$\theta(i)$ that were actually used along the simulation, must be different than
the one that was actually used at step $u$, and finally must not be equivalent
to the tile type that was used at step $u$ under rotation (if rotation is
allowed by the model in use). This number is denoted by $\alpha(i)$. We have
$\alpha(i)\ge 0$, with $\alpha(i)=0$ indicating the existence of one single
accretive sequence leading from $S$ to $\omega(i)$ and using solely tiles of the
$\theta(i)$ types used in the simulation. However, $\alpha(i)>0$ does not
indicate that the other possible sequences lead to objects that have shapes
other than that of $\omega(i)$. If $\omega(i)$ is a terminal object and has the
same shape as $\omega^*$, then we are left with a sufficient condition---i.e.,
$\alpha(i)=0$---for the termination and unicity that constraints C1 and C2
require, respectively, to hold for the tile-type set comprising the type of $S$
and the other $\theta(i)$ used in the simulation.
\end{itemize}

These are the results that get combined into fitness components for the
evaluation of $i$. In addition to them, the following further result is used in
ensuring well-formed offspring after crossover (henceforth, the set of tile
types for which an individual stands is viewed as arranged in a sequence):
\begin{itemize}
\item The active region of $i$, given by the sub-sequence delimited on the left
by the leftmost tile type actually used in assembling $\omega(i)$ and on the
right by the rightmost tile type actually used. There may be unused tile types
between the two delimiters.
\end{itemize}

\subsection{Fitness Components and Layering}

We use three fitness components in the evaluation of individual $i$, all three
real functions mapping into the interval $[0,1]$ in such a way that closer to
$1$ is better. The first one is denoted by $f(i)$ and seeks to reflect the
objective of MTSP, which is to require as small a tile-type set as possible.
This fitness component is given by
\begin{equation}
f(i)=1-\frac{1+\theta(i)}{\vert\omega(i)\vert},
\label{eq:f}
\end{equation}
where the extra $1$ in the fraction's numerator is meant to stand for the tile
type of the seed $S$. Clearly, the worst that can happen to $f(i)$ is for each
of the $\vert\omega(i)\vert$ tiles to be of a distinct type, in which case
$1+\theta(i)=\vert\omega(i)\vert$ and $f(i)=0$. Otherwise, we always have
$1+\theta(i)<\vert\omega(i)\vert$ and then $0<f(i)<1$.

The remaining two fitness components, denoted by $g(i)$ and $h(i)$, are related
to constraints C1 and C2, and therefore to ensuring that the process of
self-assembly always terminates and does so at an object of the same shape as
$\omega^*$. The fitness component $g(i)$ is given by
\begin{equation}
g(i)=\frac{2\kappa(i)}{\vert\omega(i)\vert+\vert\omega^*\vert},
\label{eq:g}
\end{equation}
where $\vert\omega^*\vert$ is the number of tiles that constitute $\omega^*$.
The first thing to notice is that, if $\omega(i)$ and $\omega^*$ have the same
shape (including coinciding hollows, if any), then $g(i)=1$, since in this case
$\kappa(i)=\vert\omega(i)\vert=\vert\omega^*\vert$. All other alternatives yield
$0<g(i)<1$, as in such cases we have $\kappa(i)<\vert\omega(i)\vert$ or
$\kappa(i)<\vert\omega^*\vert$. Additionally, if the simulations ended upon
reaching objects of the maximum allowed number of tiles, then
$\kappa(i)\le\vert\omega^*\vert<\vert\omega(i)\vert$ (assuming the simulations
are allowed to create objects with strictly more tiles than $\omega^*$, which is
certainly the case in all our experiments). In this case $g(i)<1$, thus keeping
the $g(i)=1$ goal out of reach whenever $\omega(i)$ is not a terminal object.
It then follows that $g(i)=1$ implies that $\omega(i)$ is a terminal object
having the same shape as $\omega^*$.

The third fitness component, $h(i)$, depends on the maximum number of equivalent
tiles that exist, under rotation, for the model in use. We let this be given by
the number $\rho$, which is equal to $1$ in the 2D model, to $4$ in the 2DR
model, and to $24$ in the 3DR model. This fitness component is given by
\begin{equation}
h(i)=\cases{
0,&if $\theta(i)=0$;\cr
{\displaystyle1-\frac{\alpha(i)}{\rho\vert\omega(i)\vert(1+\theta(i))}},
&if $\theta(i)>0$.\cr
}
\label{eq:h}
\end{equation}
Notice first that, if $\theta(i)=0$, then no tile has been accreted onto $S$
and we ascribe to $h(i)$ the lowest possible value. If $\theta(i)>0$, on the
other hand, then $\alpha(i)$ is defined and we use its value in obtaining
$h(i)$. Clearly, in this case $h(i)=1$ if and only if $\alpha(i)=0$, which in
turn is a sufficient (although not necessary) condition for the set comprising
the tile type of $S$ and the additional $\theta(i)$ tile types to satisfy
constraints C1 and C2, provided $\omega(i)$ is terminal and has the same shape
as $\omega^*$. Furthermore, it also holds that $0<h(i)<1$ if $\alpha(i)>0$,
since $\alpha(i)<\rho\vert\omega(i)\vert(1+\theta(i))$. Note finally that, as in
the case of $f(i)$, the $1$ added to $\theta(i)$ is meant to account for
$S$.\footnote{However, it is not in this case necessary, since the definition of
$\alpha(i)$ does not consider substitutions for $S$. Keeping the $1$ is
harmless, however, and moreover lets both $f(i)$ and $h(i)$ be expressed as
functions of the total number of tile types.}

Our three fitness components are then seen to provide some measure of coverage
of MTSP's objective and also of its constraints C1 and C2: $f(i)$ is concerned
with minimizing supply diversity while $g(i)$ and $h(i)$, when $g(i)=h(i)=1$,
ensure that constraints C1 and C2 are satisfied by the set of $1+\theta(i)$ tile
types revealed by the simulations of individual $i$. It would seem, then, that
constraint C3, the one that requires full objects to be constructed, is so far
unattended. While it is true that none of the fitness components refers to this
constraint, our reason for proceeding in this way has been both pragmatic and a
consequence of our design up to now. It has been pragmatic because we have found
a further fitness component to account for C3 to be unnecessary in our
experiments. Furthermore, and more importantly, constraint C3 is already taken
into account, albeit indirectly, in our procedure for the simulation of an
individual. This can be seen by recalling that attempts are made at filling
hollows during simulation, and that selecting the matching pair for addition to
the object at each step of the simulation privileges those pairs that will
establish the greater number of bonds, which is in accordance with what is
required by constraint C3.

It will also become apparent from the computational experiments we describe that
$f(i)$, $g(i)$, and $h(i)$ do not always vary in consonance with one another.
This has been the central motivation for us to pursue a multi-objective approach
rather than try and combine the three fitness components into one single fitness
function. What is left to specify, then, is how we use them in organizing
individuals into layers of dominance. As we mentioned earlier, we use a
Pareto-like criterion: it differs from traditional Pareto dominance in the sense
that it distinguishes $f(i)$ from both $g(i)$ and $h(i)$, since $f(i)$ is the
only one of the three that is unrelated to one of the problem's constraints,
therefore unrelated to the feasibility of individual $i$.

Our layering criteria are then the following. If $i_1$ and $i_2$ are distinct
individuals of the same population, then $i_1$ is said to dominate $i_2$ if
$i_1$ is strictly better than $i_2$ according to $g$ and no worse according to
$h$, or if $i_1$ is strictly better than $i_2$ according to $h$ and no worse
according to $g$, or yet if they are indistinguishable from each other by $g$ or
$h$ but $i_1$ is strictly better than $i_2$ according to $f$. More formally,
$i_1$ dominates $i_2$ if and only if one of the following three conditions
holds:
\begin{equation}
g(i_1)>g(i_2)\textrm{ and }h(i_1)\ge h(i_2);
\label{dominance1}
\end{equation}
\begin{equation}
g(i_1)\ge g(i_2)\textrm{ and }h(i_1)>h(i_2);
\label{dominance2}
\end{equation}
\begin{equation}
g(i_1)=g(i_2)\textrm{ and }h(i_1)=h(i_2)\textrm{ and }f(i_1)>f(i_2).
\label{dominance3}
\end{equation}
We then place strictly more weight on feasibility than on optimality.

\subsection{Crossover and Mutation}

\begin{figure}
\centering
\includegraphics[scale=0.30]{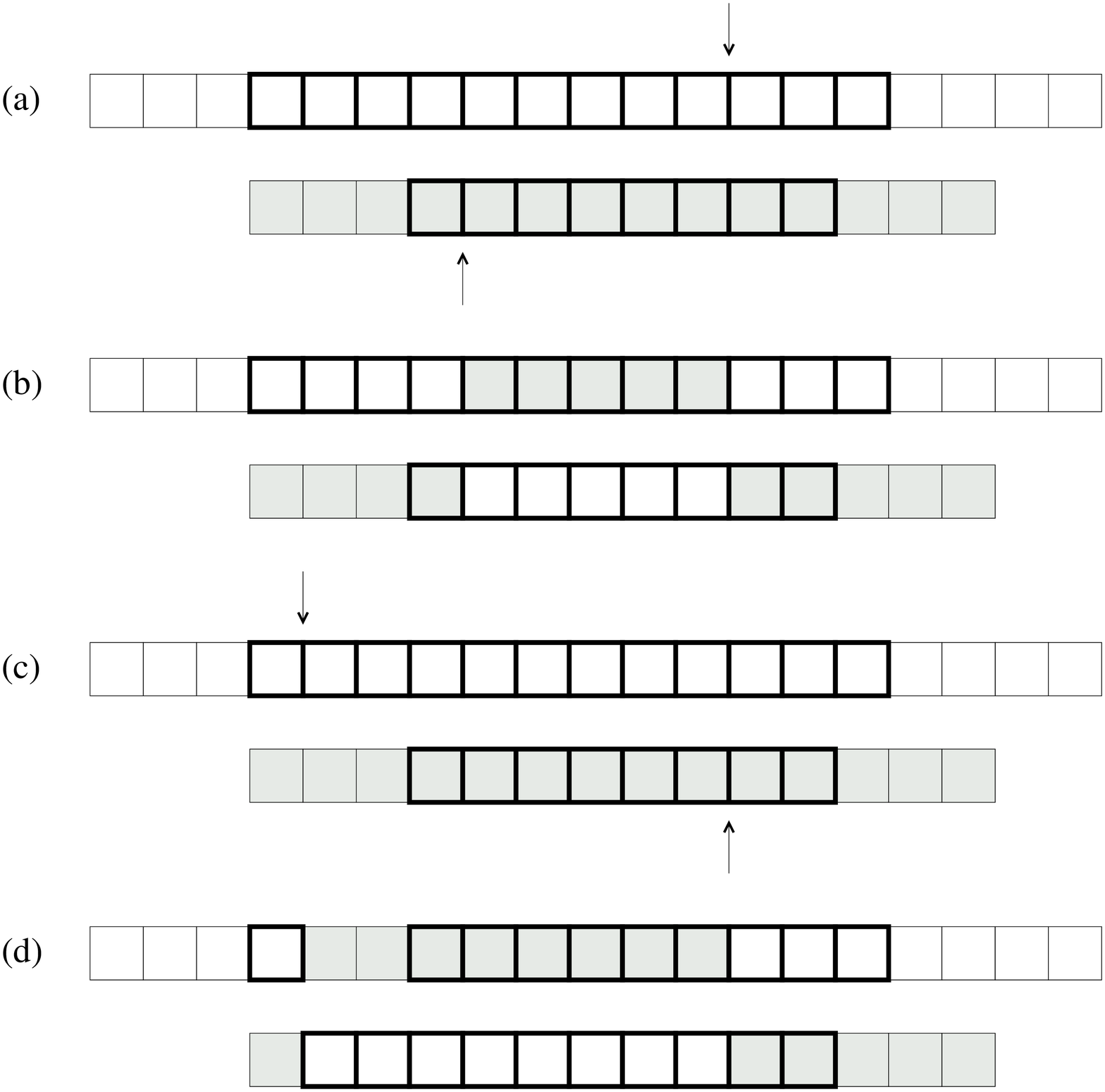}
\caption{The crossover operator, applied to the individuals in (a) to produce
those in (b), and to the ones in (c) to produce those in (d). Arrows indicate
the crossover points. Thick-line borders are used to indicate the tile types in
a parent's active region.}
\label{fig:crossover}
\end{figure}

The method we use for doing the crossover of two individuals $i_1$ and $i_2$
aims both at handling the size variability of the individuals and at ensuring
that each of the two offspring inherits material from inside the active regions
of both $i_1$ and $i_2$. The way this is achieved is by combining two well-known
techniques. First the two individuals are aligned to each other at some randomly
chosen position in such a way that the smallest one is contained inside the
other. Then two crossover points are selected randomly, provided one intersects
the active region of $i_1$ and the other that of $i_2$. Two-point crossover is
then performed as illustrated in Figure~\ref{fig:crossover}. It is simple to see
that the two resulting offspring have sizes that necessarily fall between the
allowed minimum and maximum.

Whenever it has been decided that individuals will be selected to undergo
crossover, we allow a fixed number of trials to be attempted so that the two
individuals are not closer, by Euclidean distance in the three-dimensional
space of the fitness components, than a given distance. This gives us a handle
on trying to make sure that crossover is performed on sufficiently different
individuals even when they both come from the same layer. In the event that no
pair is found with the desired characteristic, crossover is performed on the
last pair to have been selected.

Mutation is straightforward. One of the individual's tile types is selected at
random, then one of its sides also at random. The chosen side's label is then
replaced by another one from the $\tau$-dependent label table. The replacing
label is also chosen at random, the only requirement being that it differ from
the one it is replacing. If in the 2DR or 3DR model, then as mentioned earlier
a polarity is randomly chosen for the label.

\section{Computational Results}\label{sec:results}

We have conducted extensive experimentation with the evolutionary heuristic of
Section~\ref{sec:heur}, allowing for several combinations of its parameters.
Here we report on some of the most successful and, occasionally, also on some of
the experiments that, though not completely successful, have shed some light on
parameter choice in a significant way. In all experiments in two dimensions
$\omega^*$ has been an $n\times n$ square with different choices for the value
of $n$. In three dimensions, we have concentrated on the $5\times 5\times 5$
cube. Most parameter values have been the same in all experiments and appear in
Table~\ref{tab:parameters}.

\begin{table}
\centering
\caption{Common parameter values}
\label{tab:parameters}
\begin{tabular}{lc}
\hline
Parameter&Value\\
\hline
Number of generations ($G$)&$1\,000$\\
Population size&$1\,000$\\
Elitist fraction&$0.1$\\
Diversity-preserving fraction&$0.05$\\
Initial crossover probability ($p_1$)&$0.3$\\
Final crossover probability ($p_G$)&$0.7$\\
Maximum number of simulations per individual&$10$\\
Max.\ number of crossover attempts per decision to do crossover&$1\,000$\\
\hline
\end{tabular}
\end{table}

Most experiments have also used temperature $\tau=2$, a notable exception being
one of the three-dimensional experiments. The seed $S$ used in all
two-dimensional experiments has two wildcard labels placed so that $S$ is one of
the corners of the square. In three dimensions $S$ has three wildcard labels,
again intended to allow the seed to occupy one of the cube's corners.

Our results for the 2D model refer to $n=5$ or $n=15$ (but see also
Section~\ref{sec:concl}, where we discuss the $n=25$ case as well) with the
following additional parameter values. For $n=5$, a $30\times 30$ lattice,
simulated objects of no more than $100$ tiles, the size of an individual between
$25$ and $50$, and a label table with $10$ non-$\epsilon$ labels. For $n=15$,
these become $45\times 45$, $900$, $50$--$100$, and $20$, respectively.

For $n=5$ we had several successful runs, a success being characterized by the
occurrence of at least one individual $i$ for which $g(i)=h(i)=1$ (so the
$1+\theta(i)$ tile types satisfy constraints C1 and C2), and for which
constraint C3 is also satisfied, and furthermore $1+\theta(i)$, the number of
tile types (including that of $S$), is as indicated in (\ref{eq:upperbound}).
These runs correspond to the various combinations of $0$ or $0.01$ as the
minimum distance between individuals for crossover to be performed with
$W_1=W_G=15$ or $W_1=1$, $W_G=30$ (recall the latter regulate the selection of
layers through (\ref{eq:wnorm})). For the particular choice of $0$ as minimum
distance, success was achieved in all runs of a series of five with each of
$W_1=W_G=15$ or $W_1=1$, $W_G=30$. On average, this happened after about $502$
and $433$ generations, respectively. One unexpected (and, in fact, unsought for)
outcome related to one of the $W_1=W_G=15$ runs is that the resulting number of
distinct labels is $n+2=7$, better therefore than the best known estimate, which
is $n+3$ \cite{rw00}.\footnote{Minimizing the number of distinct labels is not
part of MTSP and we are therefore generally unconcerned with it in this paper.
However, this is an additional goal for which plausible arguments exist
\cite{ch03}, so our find, although serendipitous, amounts to an interesting
by-product.}

\begin{figure}[p]
\centering
\includegraphics[scale=0.30]{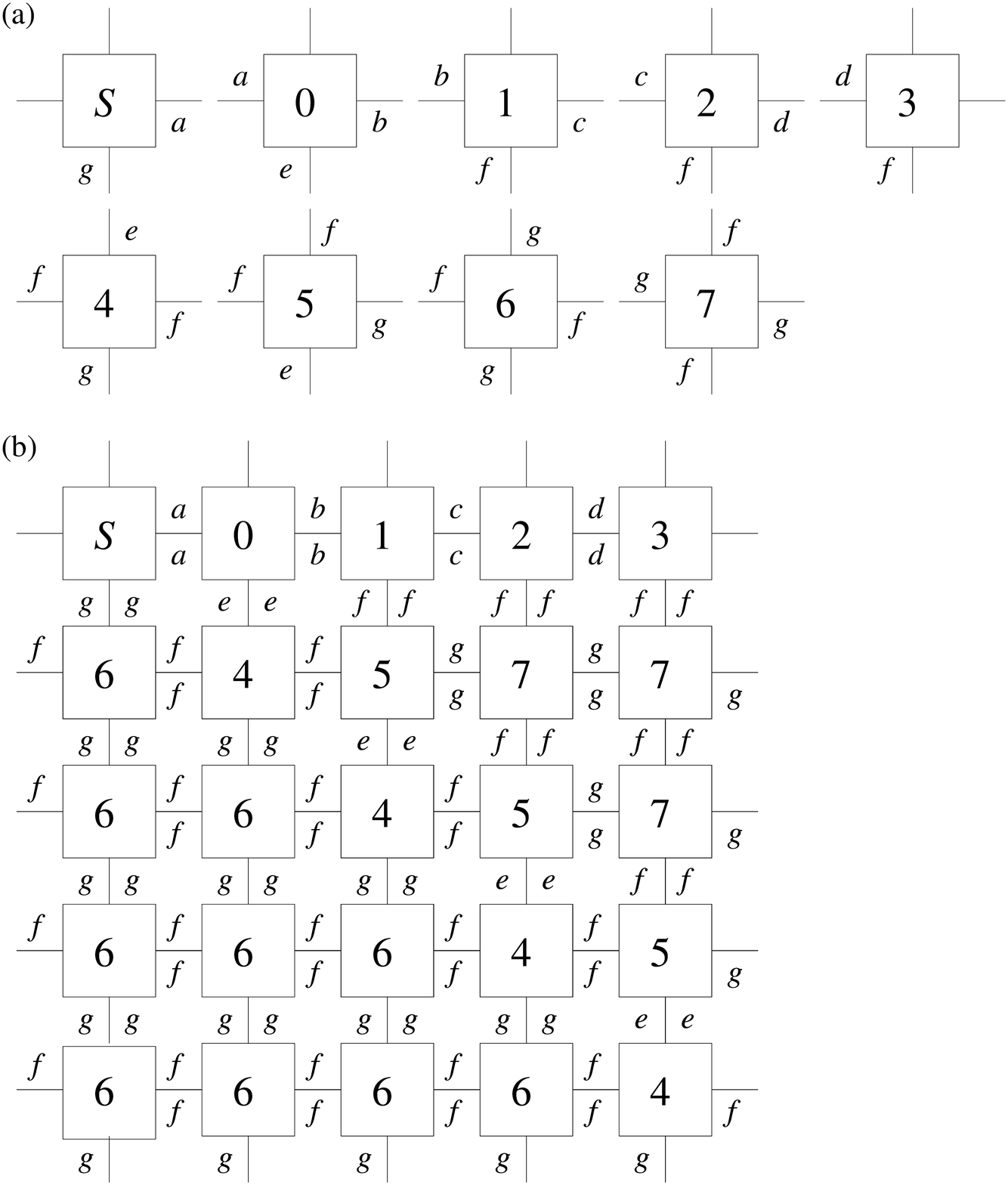}
\caption{A solution obtained in the 2D model for $n=5$. Tile types appear in
(a), the final object in (b). Temperature is $\tau=2$. Label intensities are
$I(a)=\cdots=I(e)=2$ and $I(f)=I(g)=1$.}
\label{fig:2dsolution5}
\end{figure}

\begin{figure}[p]
\centering
\includegraphics[scale=0.45]{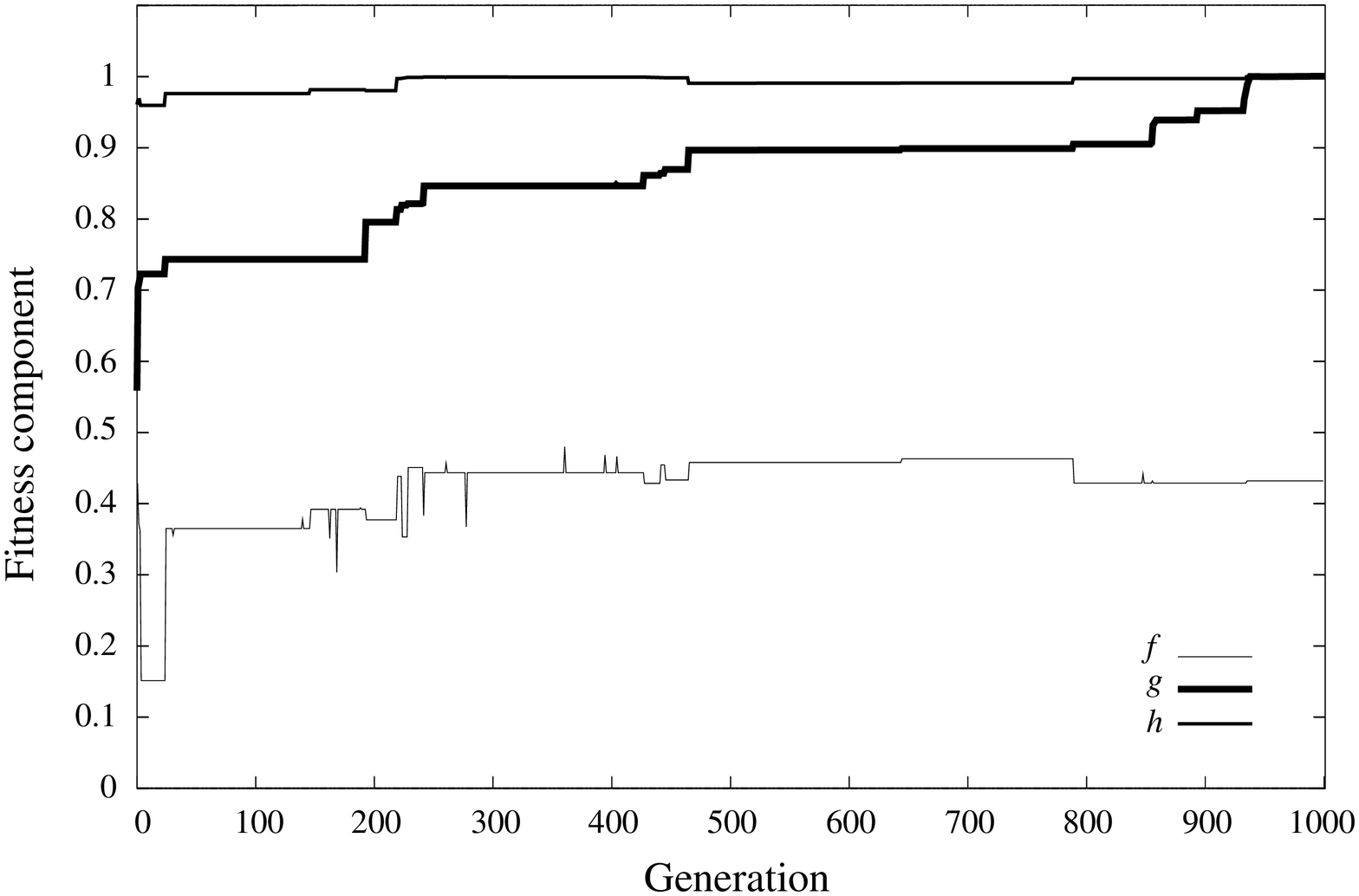}
\caption{Evolution of the fitness components for the solution shown in
Figure~\ref{fig:2dsolution5}.}
\label{fig:2dplots5}
\end{figure}

The solution achieved by this latter run is shown in
Figure~\ref{fig:2dsolution5}. The corresponding evolution plots for the three
fitness components are shown in Figure~\ref{fig:2dplots5}. These plots were
obtained according to the following (somewhat arbitrary) method, which holds
also for all other similar plots in the sequel. For each generation, the fitness
component $g$ is plotted for the best individual found so far in terms of its
$g$ value ($h$ and $f$ are used, in this order, for tie breaking). For this
best individual, $h$ and $f$ are then plotted. Notice, then, that the $g$ plot
is necessarily nondecreasing, while the plots for $h$ and $f$ may oscillate.

\begin{figure}[p]
\centering
\includegraphics[scale=0.17]{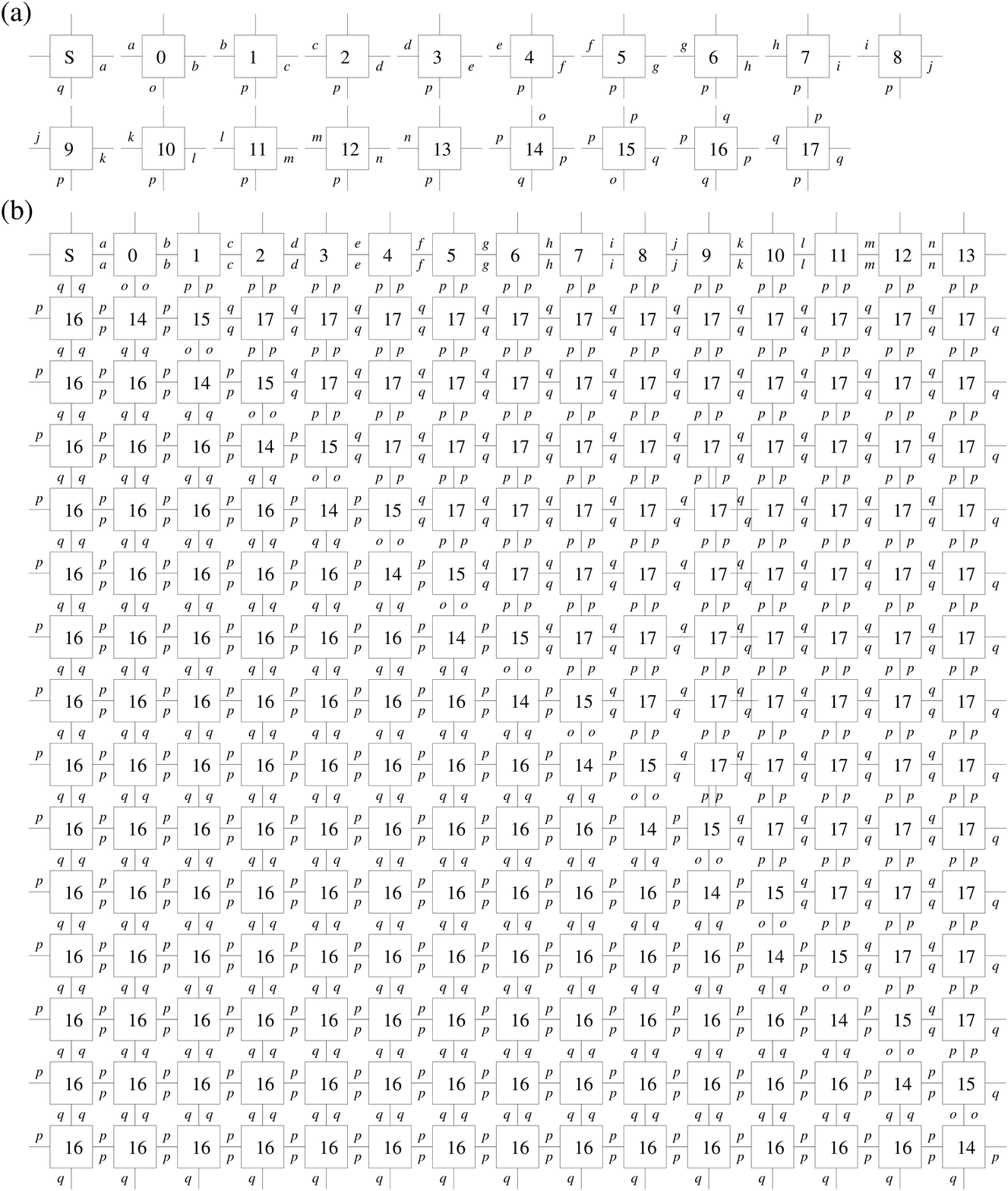}
\caption{A solution obtained in the 2D model for $n=15$. Tile types appear in
(a), the final object in (b). Temperature is $\tau=2$. Label intensities are
$I(a)=\cdots=I(o)=2$ and $I(p)=I(q)=1$.}
\label{fig:2dsolution15}
\end{figure}

\begin{figure}[p]
\centering
\includegraphics[scale=0.45]{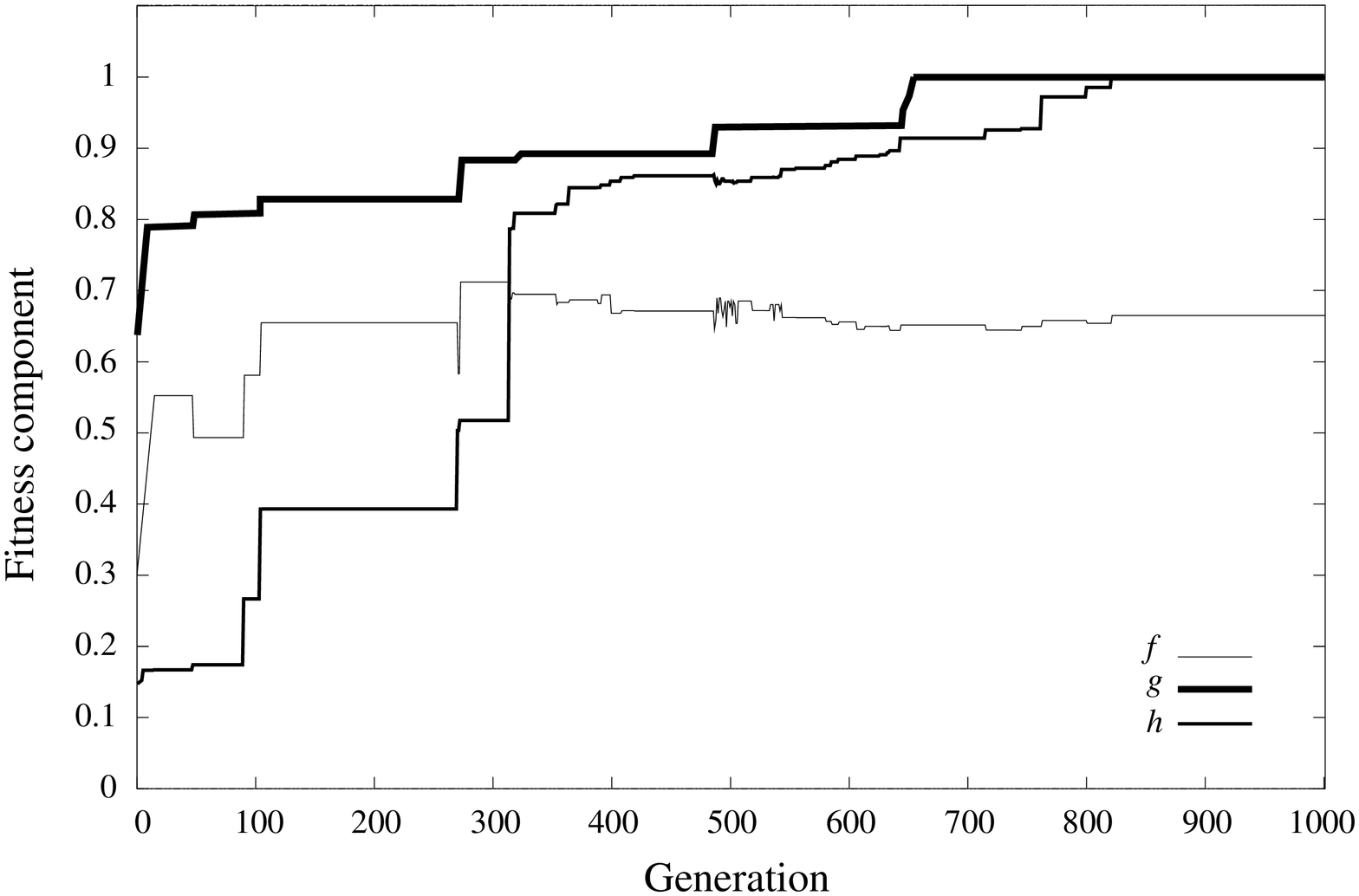}
\caption{Evolution of the fitness components for the solution shown in
Figure~\ref{fig:2dsolution15}.}
\label{fig:2dplots15}
\end{figure}

The $n=15$ case also yielded plenty of successes under the same parameter
choices described above for $n=5$ with regard to the minimum distance for
crossover and the values of $W_1$ and $W_G$. Now, however, choosing the distance
to be $0.01$ yielded success in all runs in a series of five for $W_1=W_G=15$
and another five for $W_1=1$, $W_G=30$. The solution achieved by one of the
runs with $W_1=1$, $W_G=30$ is shown in Figure~\ref{fig:2dsolution15}, with
fitness-component plots in Figure~\ref{fig:2dplots15}. When compared to the
plots of Figure~\ref{fig:2dplots5}, the ones in Figure~\ref{fig:2dplots15}
reflect the increased difficulty that comes from moving from $n=5$ to $n=15$,
since the solution is approached much more slowly for the larger square.

\begin{figure}[p]
\centering
\includegraphics[scale=0.30]{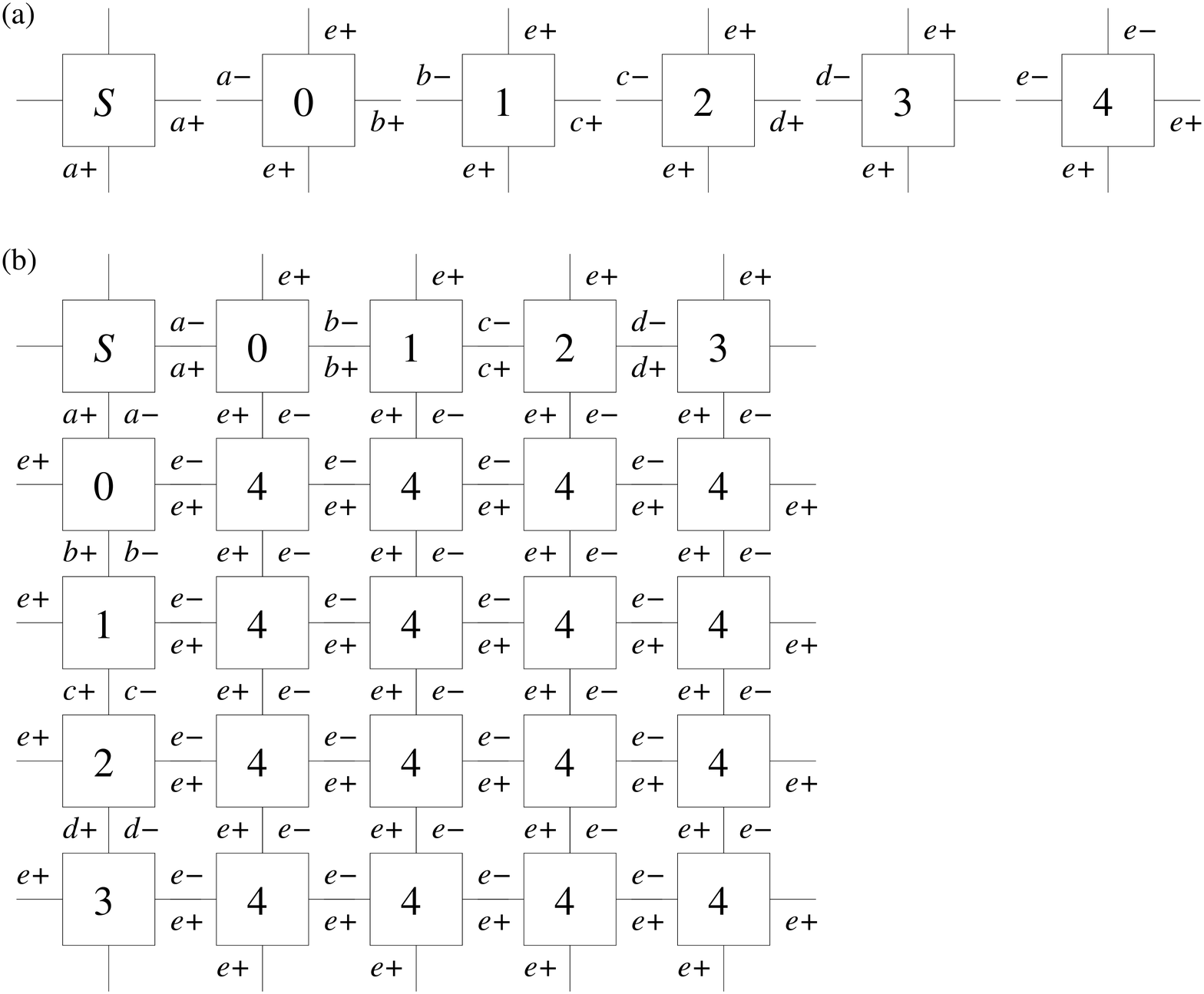}
\caption{A solution obtained in the 2DR model for $n=5$. Tile types appear in
(a), the final object in (b). Temperature is $\tau=2$. Label intensities are
$I(a)=\cdots=I(d)=2$ and $I(e)=1$.}
\label{fig:2drsolution}
\end{figure}

\begin{figure}[p]
\centering
\includegraphics[scale=0.45]{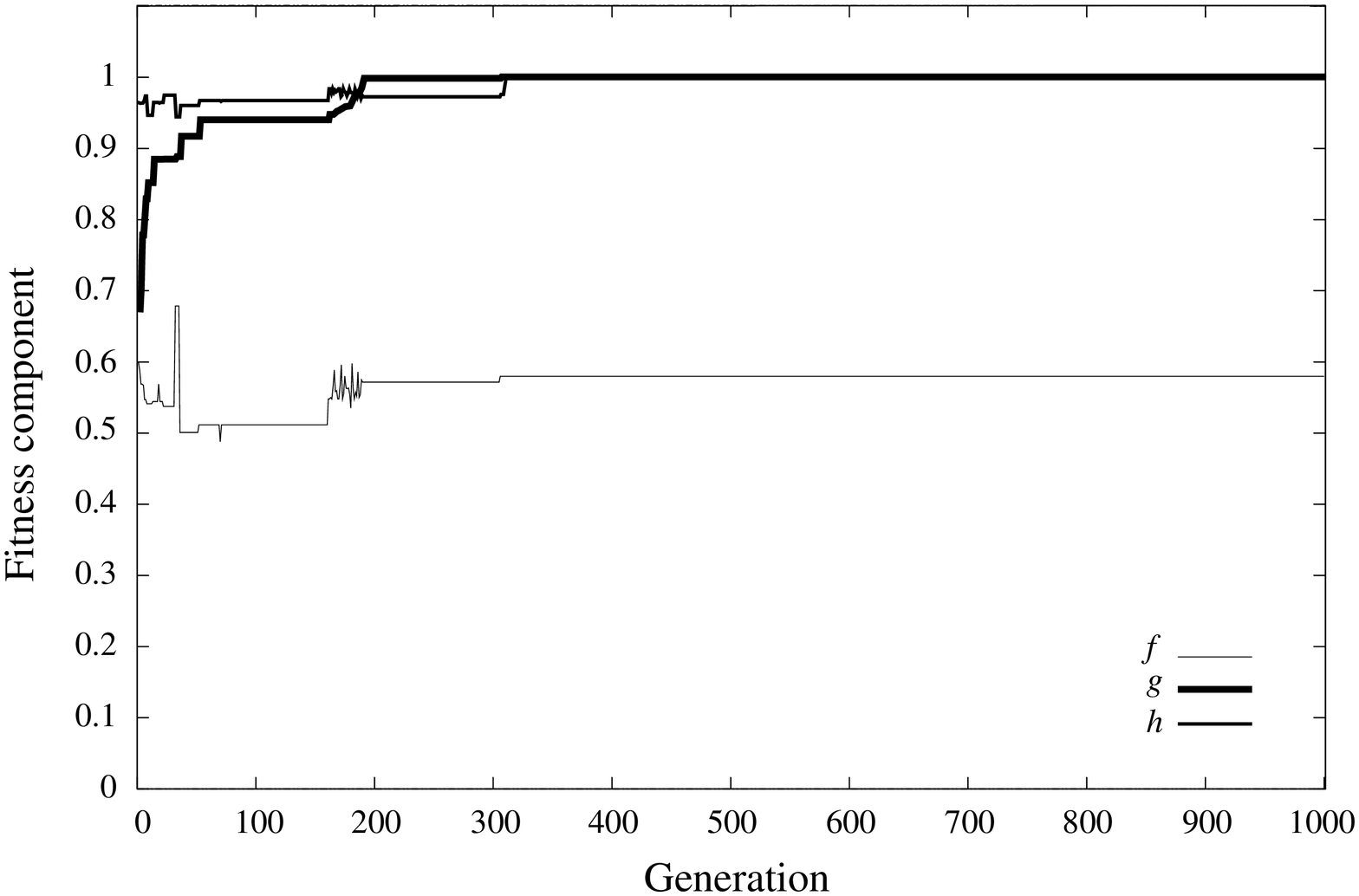}
\caption{Evolution of the fitness components for the solution shown in
Figure~\ref{fig:2drsolution}.}
\label{fig:2drplots}
\end{figure}

As we remarked earlier, as we move from the 2D model to 2DR we expect to be
afforded greater flexibility and therefore greater ease in obtaining success
with our evolutionary heuristic for MTSP. Here we illustrate this for the $n=5$
case, using the same parameters as for the case singled out above at the end of
our discussion of the 2D model. Success was obtained in all of five runs. One
of them is illustrated in Figures~\ref{fig:2drsolution} and \ref{fig:2drplots},
containing respectively the solution that was achieved and the fitness-component
plots. From the latter, in particular, the expected increased ease in finding
the solution emerges very clearly, since convergence to $1$ of both $g$ and $h$
occurs relatively early in the evolution.

It is curious to observe, in Figure~\ref{fig:2drsolution}, that the number of
tile types is $n+1=6$, therefore significantly smaller than the known upper
bound given in (\ref{eq:upperbound}) for the 2D model, which for $n=5$ is
$n+4=9$. Of course, such an upper bound may turn out to be strictly looser for
the 2DR model than it is for the 2D model once similar properties become known
for the 2DR model. Until then, we are left with this one source of assessment of
the solution shown in the figure, that is, by comparison with the 2D model.

\begin{figure}[p]
\centering
\includegraphics[scale=0.25]{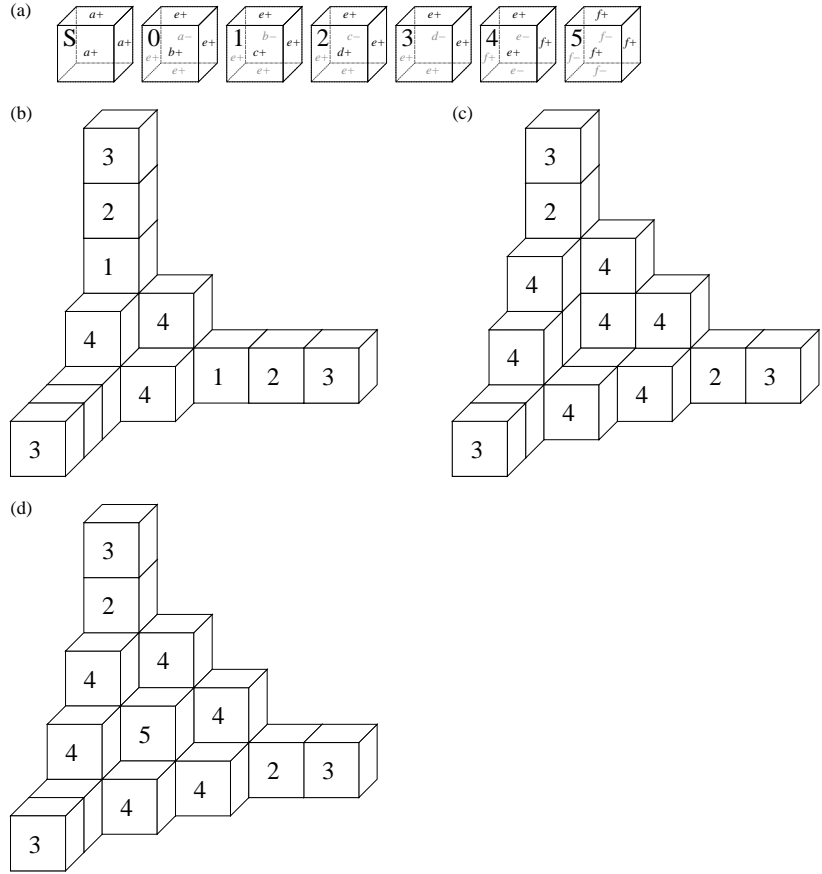}
\caption{A solution obtained in the 3DR model for $n=5$. Tile types appear in
(a), three intermediate objects on the sequence leading to the final
$5\times 5\times 5$ cube appear in (b)--(d). Temperature is $\tau=3$. Label
intensities are $I(a)=\cdots=I(d)=3$, $I(e)=2$, and $I(f)=1$. In (b), the seed
$S$ occupies the hidden corner of the cube and three tiles of type $0$ (also
hidden) are accreted onto it. Then follow tiles of types $1$, $2$, and $3$ along
each dimension. The cube is completed by filling with tiles of type $4$ the
three faces on which such tiles are already present, and finally placing
type-$5$ tiles in all remaining positions.}
\label{fig:3drsolution}
\end{figure}

\begin{figure}[p]
\centering
\includegraphics[scale=0.45]{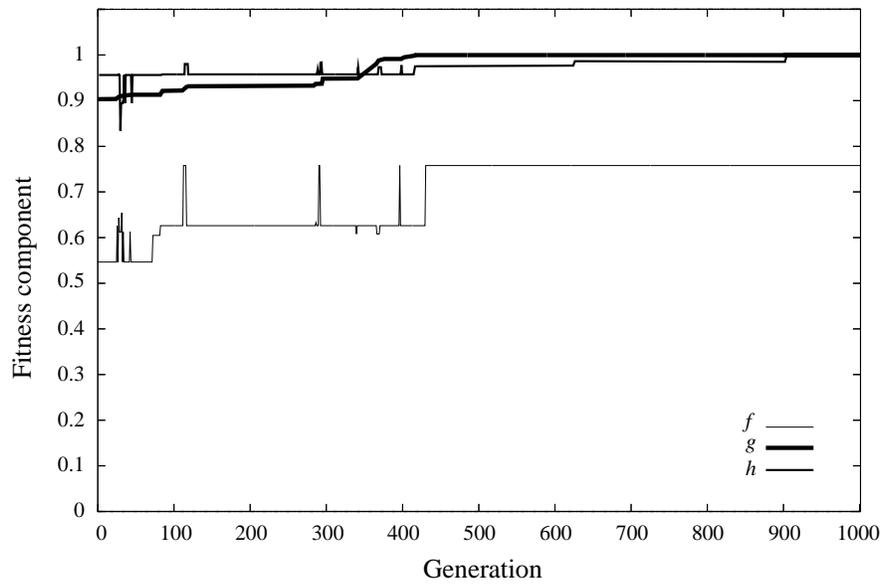}
\caption{Evolution of the fitness components for the solution shown in
Figure~\ref{fig:3drsolution}.}
\label{fig:3drplots}
\end{figure}

As we switch to three dimensions and adopt the 3DR model, solutions are still
obtained for parameter combinations similar to the ones considered so far under
two dimensions, but they no longer constitute successes in the sense we have
established. The reason for this is that, even though a full $5\times 5\times 5$
cube is obtained, its fullness depends on the order of assembly and therefore
constraint C3 is not satisfied. Our suspicions as to why this occurs fell,
naturally, on the value of $\tau$, so far kept constant at $2$. In fact,
adopting $\tau=3$ has yielded the desired solution, obtained in one single run
and shown in Figure~\ref{fig:3drsolution}. Once again, it is curious to note the
$n+2=7$ tile types used and to compare this figure to both the prediction of
(\ref{eq:upperbound}) and the results discussed heretofore. Parameter values
were the following: a $30\times 30\times 30$ lattice, simulated objects of no
more than $900$ tiles, individuals sized between $25$ and $35$, $20$
non-$\epsilon$ labels available for selection, minimum distance for crossover of
$0.01$, and finally $W_1=1$, $W_G=30$. The corresponding plots of the evolution
of the fitness components are given in Figure~\ref{fig:3drplots}.

\clearpage

\clearpage

\clearpage

\clearpage

\clearpage

\clearpage

\clearpage

\clearpage

\clearpage
\section{Discussion and Concluding Remarks}\label{sec:concl}

One practical measure of the computational hardness of solving MTSP is the time
it has taken our heuristic to complete for each of the runs described in
Section~\ref{sec:results}. Like most evolutionary algorithms, our heuristic has
a tremendous potential for parallel execution, not only because individuals may
be evaluated independently of one another, but especially because it is
precisely in computing the fitness components that lies most of the difficulty.
We ran our experiments on a small local grid with exclusive access to all
hosts, each running a Linux operating system on an Intel Pentium D (a two-core
processor) at 2.8GHz with 2GB of RAM. Average running times are given in
Table~\ref{tab:times}.

\begin{table}
\centering
\caption{Average running times}
\label{tab:times}
\begin{tabular}{lrr}
\hline
Experiment&Number of cores&Time (hours)\\
\hline
2D model, $n=5$&$8$&$4.0$\\
2D model, $n=15$&$10$&$7.7$\\
2DR model, $n=5$&$12$&$1.5$\\
3DR model, $n=5$&$12$&$12.0$\\
\hline
\end{tabular}
\end{table}

In view of the times given in the table, it comes as no surprise that our search
for appropriate parameter-value combinations has of necessity been severely
limited. This is often the case with evolutionary approaches, and no doubt the
availability of larger grids would instantly provide many fresh opportunities
for tune-up and performance studies. However, we believe to have already
provided a sort of proof of principle regarding the possibility of using an
evolutionary approach to tackle MTSP and perhaps other hard problems related to
self-assembly systems.

We find, then, that further effort will be more profitably spent if directed
toward understanding other aspects of MTSP with the use of heuristics like the
one we have introduced. We finalize by briefly reporting on one step we have
already taken in such a direction, having to do with the study of the so-called
complex seeds, i.e., seeds that, unlike the $S$ we have used throughout,
comprise more than one tile \cite{rw00}. The initial motivation came from our
heuristic's apparent inability to complete any successful runs in the 2D model
for $n=25$ using various combinations of parameter values. Interestingly,
though, once the possibility of a complex seed was brought into the scene it
became possible for our heuristic to conclude successful runs with parameter
values very similar to those used in Section~\ref{sec:results}.

\begin{figure}[p]
\centering
\includegraphics[scale=0.10]{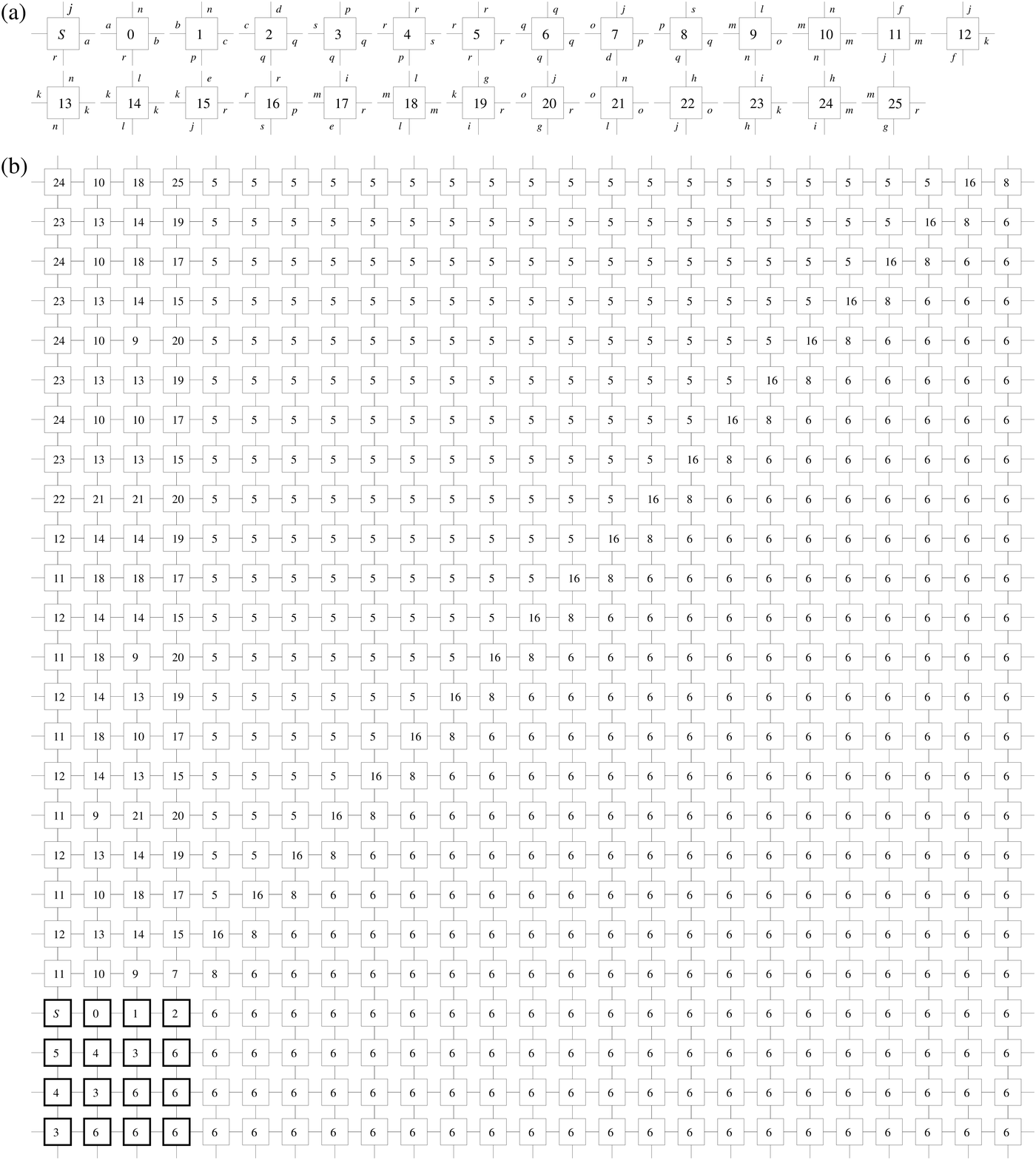}
\caption{A solution obtained in the 2D model for $n=25$. Tile types appear in
(a), the final object in (b). The complex seed used is shown in part (b)
comprising all sixteen tiles with thick-line borders. Temperature is $\tau=2$.
Label intensities are $I(a)=\cdots=I(h)=I(p)=2$ and
$I(i)=\cdots=I(o)=I(q)=I(r)=1$.}
\label{fig:2dsolution25}
\end{figure}

\begin{figure}[p]
\centering
\includegraphics[scale=0.45]{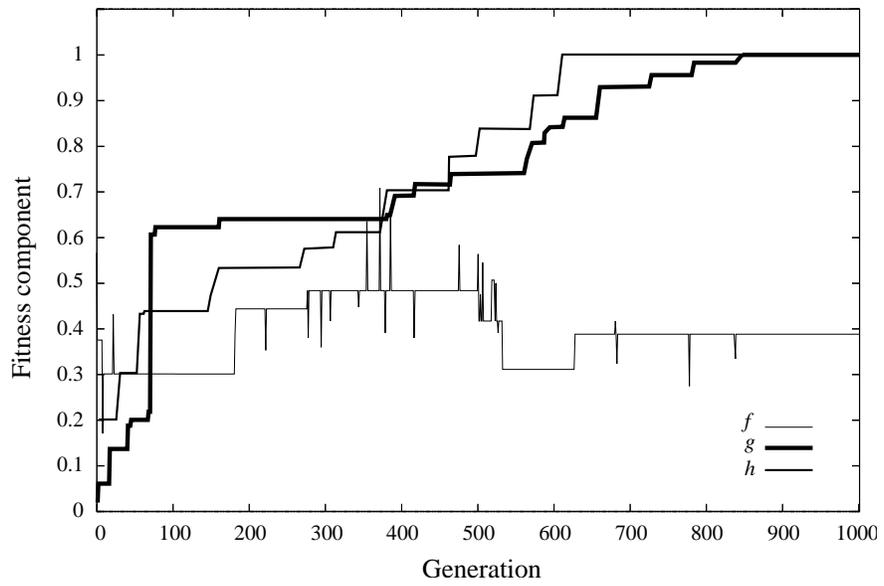}
\caption{Evolution of the fitness components for the solution shown in
Figure~\ref{fig:2dsolution25}.}
\label{fig:2dplots25}
\end{figure}

The result for one of the successful runs is shown in
Figure~\ref{fig:2dsolution25}, where the solution is given along with the
complex seed that was used,\footnote{This seed is based on the solution to MTSP
for $n=4$, for which (\ref{eq:upperbound}) predicts as an upper bound exactly
the $n+4=8$ tile types that it contains. It has wildcard labels on all tile
sides that face inward with respect to the final square.} and in
Figure~\ref{fig:2dplots25}, which contains the fitness-component plots. Even
with the use of the complex seed, though, comparing these plots with those of
Figure~\ref{fig:2dplots15}, relative to the $n=15$ case in the 2D model, reveals
the increased hardness due to the greater value of $n$, as convergence to $1$ of
the $g$ and $h$ fitness components occurs more slowly. For this run with $n=25$
we used the same basic set of parameters as in Section~\ref{sec:results} for the
2D cases, with the following differences/enlargements: fixed crossover
probability of $p_1=p_G=0.3$, a $100\times 100$ lattice, simulated objects of no
more than $2\,500$ tiles, individuals sized between $100$ and $150$, $20$
non-$\epsilon$ labels available for selection, minimum distance for crossover of
$0.01$, and finally $W_1=1$, $W_G=30$. The average running time was of $10$
hours on $10$ cores.

\clearpage

\clearpage

\clearpage
\subsection*{Acknowledgments}

The authors acknowledge partial support from CNPq, CAPES, and a FAPERJ BBP
grant.

\bibliography{selfassembly}
\bibliographystyle{plain}

\end{document}